\documentclass[12 pt]{article}
\usepackage{authblk}
\usepackage{hyperref}
\usepackage{graphicx}
\usepackage{float}
\usepackage{amsmath}
\usepackage{amssymb}
\usepackage{subcaption}
\usepackage{float}

\usepackage[verbose=true,letterpaper]{geometry}
\AtBeginDocument{
	\newgeometry{
		textheight=9in,
		textwidth=5.5in,
		top=1in,
		headheight=12pt,
		headsep=25pt,
		footskip=30pt
	}
	
}

\usepackage[sorting=none, backend=biber]{biblatex}
\title{Variational Autoencoder Layer}
\author{Gananath R\\
	\texttt{gananathr@gmail.com}
}

\date{}

\begin{document}
	
\maketitle

\begin{abstract}
	\noindent Variational Autoencoders (VAEs) belong to a family of autoencoders with probabilistic properties, making them well suited for generating data by producing a smooth and continuous latent space. Despite being introduced over a decade ago, the method continues to be widely adopted in both research and industry for diverse applications. While VAEs are typically used as standalone models, this paper introduces a novel approach to integrate them as a neural network layer. Furthermore, a new training strategy is proposed for models incorporating these layers, and their performance is thoroughly analyzed.
\end{abstract}

\section{Introduction}
Autoencoder (AE) is a special class of neural networks widely used in deep learning (DL). They are generally trained without labeled data, in a unsupervised manner, where the model learns to reconstruct its own inputs. An AE comprises two parts: an encoder and a decoder. The encoder compresses and encodes the input data into a meaningful low-dimensional representation, while the decoder reconstructs the input from this latent representation. In modern AEs, backpropagation and reconstruction loss are important components of DL training. AEs have many applications, with prominent use in dimensionality reduction and feature learning. Since their introduction nearly forty years ago, various variants of AEs have been proposed, including sparse, denoising, and contractive AEs. VAEs are a special flavor of autoencoders used not only for representation learning but also for data generation \cite{Bank2020, Mienye2025, Tschannen2018}.

VAEs fall under the class of generative autoencoders, which learn data distributions and can synthesize new samples. They were first proposed in 2013 by Kingma and Welling as an extension inspired by the Helmholtz Machine. The main difference between a vanilla AE and a VAE is that a vanilla AE learns fixed low-dimensional latent representations of the input, whereas a VAE learns the underlying probability distribution of the input data. A VAE can be viewed as a deep latent variable model (DLVM) that assumes the observed data $\mathbf{x}$ is generated from a latent variable $\mathbf{z}$ through an underlying probabilistic process.

\begin{equation*}
p_{\theta}(\mathbf{x}, \mathbf{z}) = p_{\theta}(\mathbf{x} \mid \mathbf{z}) p(\mathbf{z})
\end{equation*}

The main challenge in DLVM is that exact posterior inference is generally intractable because computing the marginal likelihood $p_{\theta}(\mathbf{x})$ requires integrating over all latent variables. To address this, variational inference is employed by introducing a parametric approximate posterior $q_{\phi}(\mathbf{z} \mid \mathbf{x})$, also known as an encoder or recognition model, which serves as a tractable approximation to the true posterior $p_{\theta}(\mathbf{z} \mid \mathbf{x})$, thereby enabling optimization of the model through a lower bound on the marginal likelihood \cite{Kingma2013, Kingma2019}.

Backpropagation is a cornerstone of training in modern artificial neural networks, yet it lacks a biological counterpart in natural brains. As an alternative, Hinton introduced the Forward-Forward algorithm, which diverges from traditional backpropagation. Building on this, researchers developed Forward-Forward Contrastive Learning (FFCL), a three-stage method. However, FFCL still depended on backpropagation in its final stage. To address this limitation, Improved Forward-Forward Contrastive Learning (IFFCL) was proposed; a streamlined, single-stage approach that employs two distinct neural networks. In IFFCL, these networks are trained layer-wise using a shared loss function for localized updates. The output of one model serves as a reference for training the corresponding layer in the other network. \cite{Gananath2024}. The IFFCL framework forms the foundation of the training procedure employed in this research. 

\section{Methods}

\begin{figure}[t]
	\centering
	\includegraphics[scale=0.35]{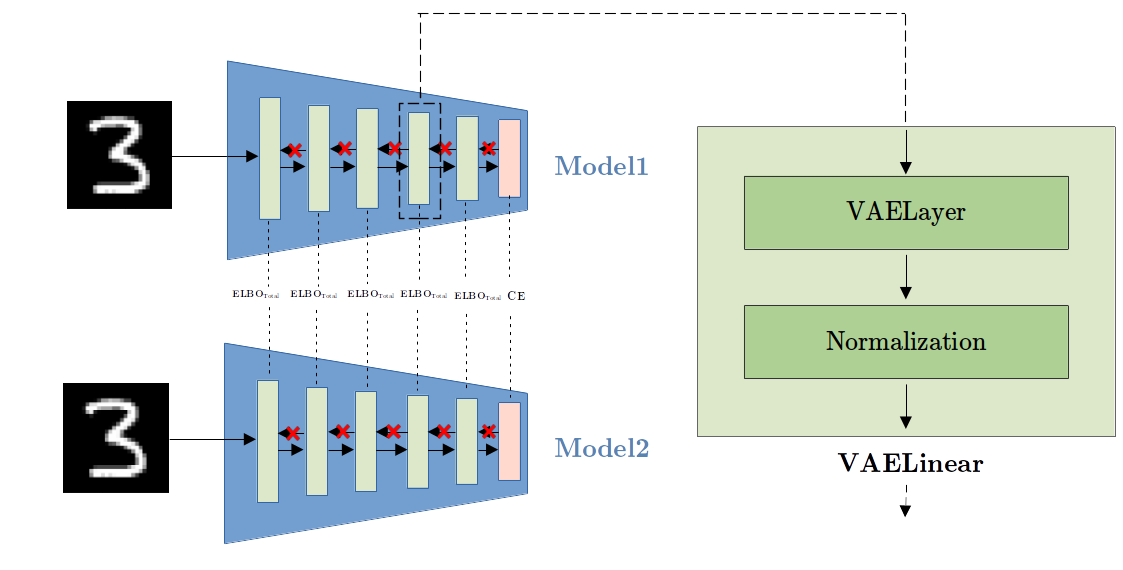}
	\caption{This figure provides a visual overview of our IFFCL model, including a detailed, magnified depiction of the VAELinear layer and its constituent components.}
	\label{fig:fullmodel}
\end{figure}

Traditionally, VAEs are trained as full networks, consisting of multiple interconnected layers much like AEs, which similarly feature encoder and decoder sections with equivalent functions. However, in this paper, we shift our approach by treating VAEs not as standalone models but as individual neural network layers. Additionally, by incorporating the IFFCL algorithm, we introduce a training method that eliminates the need for backpropagation through layers.

\subsection{VAELinear}

A standard hidden linear (dense) layer in a multilayer perceptron performs a linear transformation of the input, followed by a nonlinear activation function. The purpose of this transformation is to map the input data into a new representation that can be further processed by subsequent layers.

Given an input vector $\mathbf{x} \in \mathbb{R}^{d_{\mathrm{in}}}$, the layer computes a weighted sum using a weight matrix $W \in \mathbb{R}^{d_{\mathrm{out}} \times d_{\mathrm{in}}}$ and a bias vector $\mathbf{b} \in \mathbb{R}^{d_{\mathrm{out}}}$. The resulting output vector $\mathbf{z} \in \mathbb{R}^{d_{\mathrm{out}}}$ is given by

\begin{equation*}
	\mathbf{z} = W\mathbf{x} + \mathbf{b}
\end{equation*}

The transformed representation $z$ is then passed through a nonlinear activation function before being forwarded to the next layer.

Our VAELinear architecture comprises two key components: a VAELayer and a normalization layer. The VAELayer operates conceptually and functionally like a standard VAE model, with the primary distinction being its implementation as a neural network layer rather than a standalone model. Unlike conventional Linear/Dense layers that rely on standard linear transformations, we employ a specialized linear transformation introduced in the original VAE paper called the reparameterization trick.

\begin{equation*}
	\begin{aligned}
		\boldsymbol{\mu}
		&= W_1\mathbf{x} \\
		\boldsymbol{\sigma}
		&= \exp\!\left(\frac{1}{2}\log \boldsymbol{\sigma}^{2}\right) \\		
		log \boldsymbol{\sigma}^{2}
		&= W_2\mathbf{x} \\
		\mathbf{z}
		&= \boldsymbol{\mu}
		+ \boldsymbol{\sigma} \odot \boldsymbol{\varepsilon} + \mathbf{b},
		\qquad
		\boldsymbol{\varepsilon} \sim \mathcal{N}(\mathbf{0}, I)
	\end{aligned}
\end{equation*}

The mean $\boldsymbol{\mu}$ and standard deviation $\boldsymbol{\sigma}$ are trainable functions with their own parameters. The reparameterization trick was first proposed as a deterministic alternative for the stochastic sampling process so that gradients can pass through it while training.

In addition to the VAELayer, VAELinear also consists of a normalization layer. In this paper we have used batch normalization but in our experiments layer normalization also gives similar results. Figure \ref{fig:fullmodel} is our IFFCL implementation with VAELayer.

\subsection{MVAE}
The original VAE uses the Evidence Lower Bound (ELBO) loss as the fundamental objective function for variational inference. The primary goal is to maximize the ELBO, which in turn maximizes a lower bound on the log-likelihood, $\log p_{\theta}(\mathbf{x})$, of the observed data. The first term in the ELBO corresponds to the reconstruction loss, while the second term represents the Kullback--Leibler (KL) divergence.

\begin{equation*}
	\text{ELBO}(\mathbf{x})
	=
	\mathbb{E}_{q_{\phi}(\mathbf{z}\mid\mathbf{x})}
	\left[
	\log p_{\theta}(\mathbf{x}\mid\mathbf{z})
	\right]
	-
	\mathrm{KL}
	\left(
	q_{\phi}(\mathbf{z}\mid\mathbf{x})
	\,\|\, 
	p(\mathbf{z})
	\right)
\end{equation*}

The conventional ELBO formulation is designed under the assumption that training is performed on a single-modality dataset. In contrast, the IFFCL framework employs at least two models during the training process. In this work, we consider the input to the second model as an additional modality associated with the input of the first model. Under this multimodal setting, the standard ELBO objective is no longer suitable and therefore cannot be directly applied.

To address multimodal learning, Wu and Goodman introduced the Multimodal Variational Autoencoder (MVAE) in 2018, together with a subsampled training strategy for variational autoencoders. One of the primary strengths of MVAE is its ability to learn a unified latent representation that captures the joint distribution of multiple modalities. The following equation presents the joint ELBO objective for a bimodal MVAE.

\begin{equation*}
	\mathrm{ELBO}(\mathbf{x}_1, \mathbf{x}_2)
	=
	\mathbb{E}_{q_{\phi}(\mathbf{z}\mid \mathbf{x}_1,\mathbf{x}_2)}
	\left[
	\log p_{\theta}(\mathbf{x}_1\mid \mathbf{z})
	+
	\log p_{\theta}(\mathbf{x}_2\mid \mathbf{z})
	\right]
	-
	\mathrm{KL}\!\left(
	q_{\phi}(\mathbf{z}\mid \mathbf{x}_1,\mathbf{x}_2)
	\,\|\, 
	p(\mathbf{z})
	\right)
\end{equation*}

where the posterior is defined using a product of experts (PoE) formulation:

\begin{equation*}
	q_{\phi}(\mathbf{z} \mid \mathbf{x}_1, \mathbf{x}_2)
	\propto
	p(\mathbf{z})\,
	q_{\phi_1}(\mathbf{z} \mid \mathbf{x}_1)\,
	q_{\phi_2}(\mathbf{z} \mid \mathbf{x}_2)
\end{equation*}

The joint ELBO loss alone is not sufficient as the final objective for training a multimodal network. In practice, the MVAE objective optimizes ELBOs over different subsets of modalities, ensuring consistent inference and robustness to missing modalities during training and evaluation \cite{Wu2018}. In this paper, we use a simplified version of the MVAE objective for computing the total loss, instead of the original subset-based ELBO formulation proposed in the original paper.

\begin{equation*}
	\mathrm{ELBO}_{\mathrm{Total}}
	=
	\mathrm{ELBO}(\mathbf{x}_1, \mathbf{x}_2)
	+
	\mathrm{ELBO}(\mathbf{x}_1)
	+
	\mathrm{ELBO}(\mathbf{x}_2)
\end{equation*}

\subsection{Reconstruction Loss}
In our implementation, the VAE is employed as a layer and does not include an explicit decoder network, unlike the standard formulation. Instead, the decoder is assumed to be an identity mapping, such that the reconstructed data are identical to the latent representation. Consequently, the reconstructed data from the first model's layer are denoted by $\mathbf{x} \equiv \mathbf{z}$, and the reconstructed data from the second model's corresponding layer are denoted by $\mathbf{\hat{x}} \equiv \mathbf{z}$, where both share the same latent space representation. Under the Gaussian prior imposed on $\mathbf{z}$, the latent representation is encouraged to follow a Gaussian distribution during training. Since the decoder is assumed to be an identity mapping, the reconstruction $\mathbf{x}$ and $\mathbf{\hat{x}}$ inherits the same distributional characteristics as $\mathbf{z}$.

In the MVAE paradigm, the approximate posterior is defined as $q_{\phi}(\mathbf{z} \mid \mathbf{x}, \mathbf{\hat{x}})$, which represents the distribution of the latent variable $\mathbf{z}$ conditioned on the observations from both modalities, $\mathbf{x}$ and $\mathbf{\hat{x}}$. By combining this with the IFFCL concept, we assume that the conditional likelihood of a layer can be represented as an isotropic Gaussian distribution whose mean is given by the latent variable $\mathbf{z}$. Since $\mathbf{\hat{x}} \equiv \mathbf{z}$, the conditional likelihood is given by

\begin{equation*}
	p_{\theta}(\mathbf{x}\mid\mathbf{z})
	=
	\mathcal{N}(\mathbf{x}; \mathbf{\hat{x}}, \sigma^2 I).
\end{equation*}

Under these assumptions, the reconstruction loss can be derived from the negative log-likelihood, where the expectation is taken over the approximate posterior, averaging the reconstruction error across latent samples.

\begin{equation*}
	\begin{aligned}
		\log p_{\theta}(\mathbf{x}\mid \mathbf{z})
		&=
		-\frac{D}{2}\log(2\pi\sigma^2)
		-\frac{1}{2\sigma^2}\|\mathbf{x}-\mathbf{\hat{x}}\|^2
		\\
		\mathcal{L}_{\mathrm{recon}}
		&=
		-\mathbb{E}_{q_{\phi}(\mathbf{z}\mid \mathbf{x},\mathbf{\hat{x}})}
		\left[
		\log p_{\theta}(\mathbf{x}\mid \mathbf{z})
		\right]
		\\
		&=
		\frac{1}{2\sigma^2}
		\mathbb{E}_{q_{\phi}(\mathbf{z}\mid \mathbf{x},\mathbf{\hat{x}})}
		\left[
		\|\mathbf{x}-\mathbf{\hat{x}}\|^2
		\right]
		+\text{const}
		\\
		&=
		\lambda
		\mathbb{E}_{q_{\phi}(\mathbf{z}\mid \mathbf{x},\mathbf{\hat{x}})}
		\left[
		\|\mathbf{x}-\mathbf{\hat{x}}\|^2
		\right]
		+\text{const},
		\qquad
		\lambda=\frac{1}{2\sigma^2}.
		\\
		&\approx
		\lambda
		\|\mathbf{x}-\mathbf{\hat{x}}\|^2
		+\text{const}.
	\end{aligned}
\end{equation*}

where $\sigma^2 > 0$ denotes the variance of the Gaussian likelihood. In the limit $\sigma^2 \to 0$, the distribution converges to a Dirac delta function, yielding a deterministic mapping. A strictly positive variance is required to define a proper stochastic likelihood in the VAE structure. For ease of implementation, we set $\lambda = 1$ in all experiments.

\section{Results and Discussion}

\subsection{MNIST Dataset}
\begin{figure}[h]
	\centering
	\includegraphics[scale=0.48]{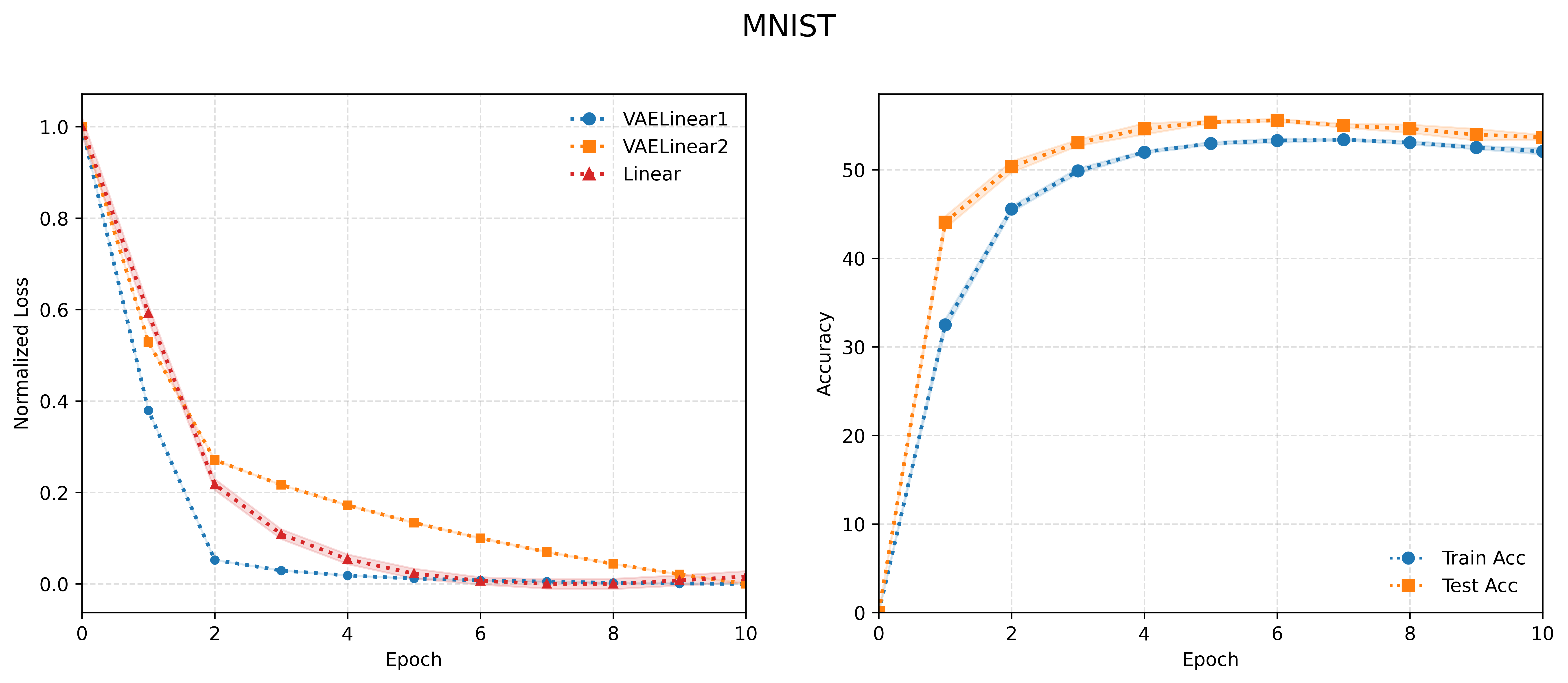}
	\caption{The figure illustrates the triplicate experimental results of MNIST dataset, where the mean is shown as a dotted curve and the standard deviation is represented by shaded regions. (Left) Results of normalized training loss across all three layers; (Right) Plot of training and testing accuracy for a single model.}
	\label{fig:result_mnist}
\end{figure}

The training followed the same procedure used in the original $\text{IFFCL}$ study. The model contains of three layers: the first is VAELinear1, implemented as $\textit{VAELinear}(784, 512)$, where 784 is the input dimension and 512 is the output dimension; the second is VAELinear2, implemented as $\textit{VAELinear}(512, 256)$; and the final layer is a standard $\textit{Linear}(256, 10)$. Each layer was trained independently, with no backpropagation across layers. We used the Adam optimizer for all layers with a fixed learning rate of 0.0001 and L2 regularization with a weight decay of 0.0001. All layers used $\tanh$ activations after their outputs, except for the final classification linear layer, which used a softmax activation. The models were trained for 10 epochs on the MNIST training set using positive sampling and evaluated on the MNIST test set. The VAE layers were trained with the $\mathrm{ELBO}_{\mathrm{Total}}$ objective, while the classification layer was trained with cross-entropy loss. To improve reliability, all experiments were repeated three times.

Figure \ref{fig:result_mnist} presents the training results. The left panel shows the normalized training loss for each layer, while the right panel displays the accuracy curves for both the training and evaluation datasets. All three layers exhibit a sharp decrease in loss, which indicates effective model training. The two VAE-based intermediate layers, VAELinear1 and VAELinear2, show very little variation in loss, whereas the final layer shows a slight fluctuation. The accuracy curves increase steadily with each epoch, reaching peak accuracies of 53\% on the training set and 55\% on the test set.

We observed that, with longer training, both training and testing accuracies gradually decrease. Closer examination indicates that the VAE-based layers remain stable, whereas the final classification layer degrades and loses its generalization ability. As shown in Figure \ref{fig:result_mnist}, after epoch 6 the accuracy begins to drop steadily, while the loss of the Linear layer starts increasing. We also performed similar experiments on a different, smaller dataset, where no decrease in training or testing accuracy was observed even after tens of thousands of training epochs (data not included).

\subsection{Fashion MNIST Dataset}

\begin{figure}[h]
	\centering
	\includegraphics[scale=0.48]{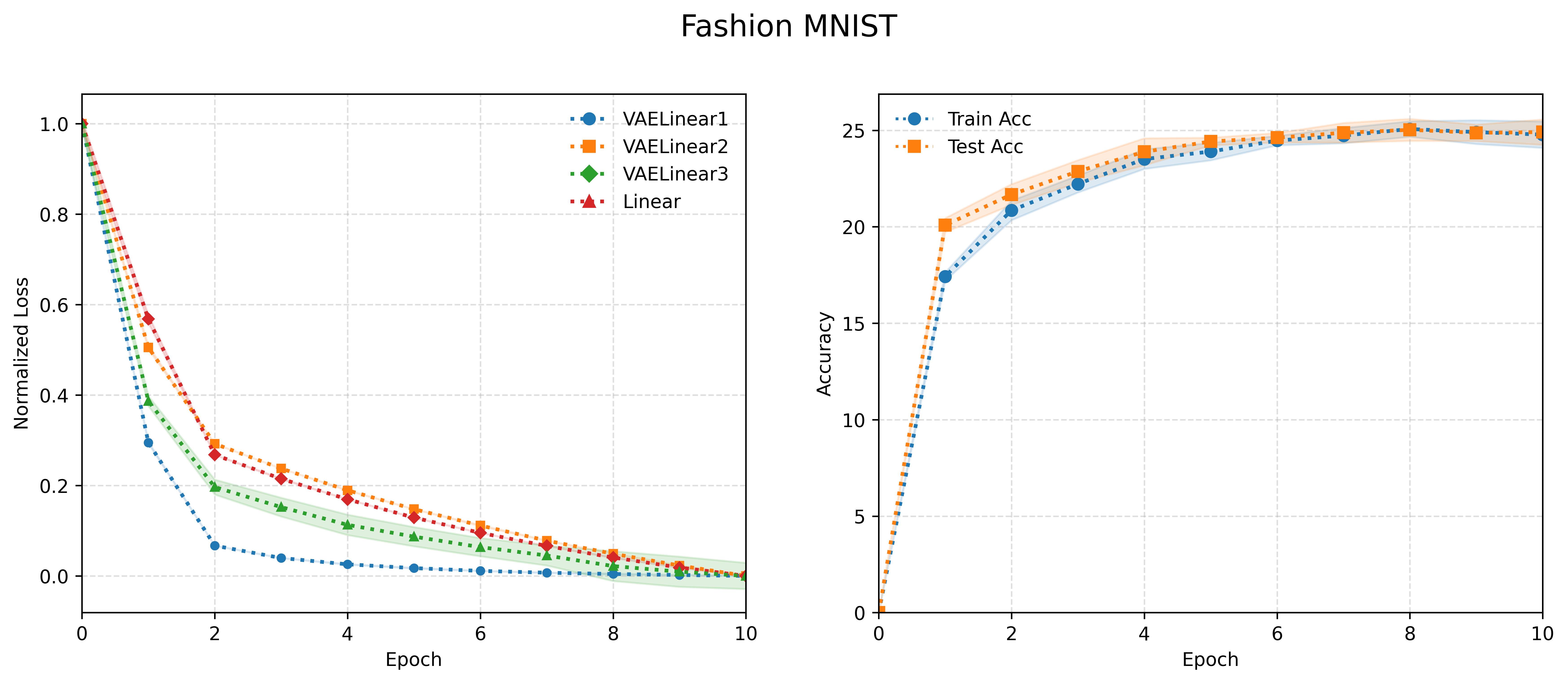}
	\caption{The figure illustrates the triplicate experimental results of Fashion MNIST dataset, where the mean is shown as a dotted curve and the standard deviation is represented by shaded regions. (Left) Results of normalized training loss across all four layers; (Right) Plot of training and testing accuracy for a single model.}
	\label{fig:result_fmnist}
\end{figure}

In the Fashion-MNIST experiments, only minor changes were made to the model architecture, while the hyperparameters and loss functions for the corresponding layers were kept identical to those used in the MNIST experiments. The primary changes involved the network depth and the number of neurons in each layer. The resulting architecture consists of four layers: three \textit{VAELinear} layers followed by a final \textit{Linear} classification layer. From the input to the output, the layers are defined as \textit{VAELinear}(784, 256), \textit{VAELinear}(256, 128), \textit{VAELinear}(128, 64), and \textit{Linear}(64, 10). These layers are referred to as \textit{VAELinear1}, \textit{VAELinear2}, \textit{VAELinear3}, and \textit{Linear}, respectively.

Similar to the results observed on the MNIST dataset, the loss curves across all layers decreased steadily during training, while the accuracy increased in the initial epochs and eventually plateaued toward the end, as shown in Figure~\ref{fig:result_fmnist}. Only marginal improvements were observed after 10 epochs when compared to MNIST. This behavior can be attributed to the higher visual complexity of the Fashion-MNIST dataset, which contains more diverse shapes and textures compared to the relatively simple digit images in MNIST.

\subsection{SVHN Dataset}
\begin{figure}[h]
	\centering
	\includegraphics[scale=0.48]{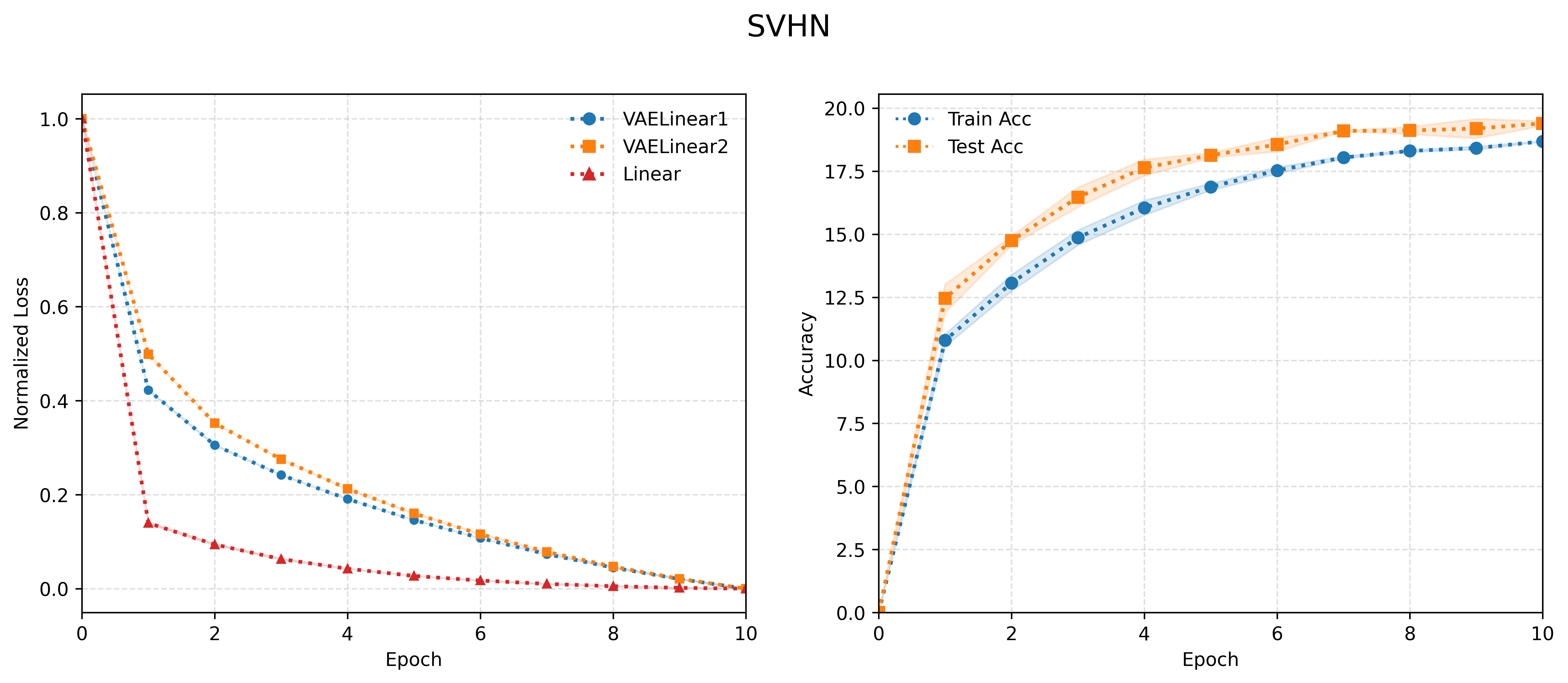}
	\caption{The figure illustrates the triplicate experimental results of SVHN dataset, where the mean is shown as a dotted curve and the standard deviation is represented by shaded regions. (Left) Results of normalized training loss across all three layers; (Right) Plot of training and testing accuracy for a single model.}
	\label{fig:result_svhn}
\end{figure}

The same model architecture and training protocol used for the MNIST dataset were applied to the SVHN dataset. The only modification was in the input dimensionality, where the network input was changed from 784 features to 3072 features obtained by flattening $32 \times 32 \times 3$ RGB images. The training behavior exhibits a trend similar to that observed in the previous datasets, Figure~\ref{fig:result_svhn}. The Street View House Numbers (SVHN) dataset is unique in our experiments, as it is the only dataset consisting of RGB images, in contrast to the grayscale images used in the earlier datasets.The limited gain in accuracy is likely due to the richer feature representation provided by the RGB channels compared to grayscale images, together with the intrinsic properties of the SVHN dataset.

For a comprehensive study, we trained and evaluated our models on three diverse datasets. The \textit{VAELinear} layer requires approximately twice the computational cost compared to a standard \textit{Linear} layer. Due to limited computational resources, we constrained the number of layers and training epochs and used a batch size of 64, balancing computational efficiency with model performance. The relatively low accuracy observed across all datasets can be attributed to the shallow network architecture used for training, as well as the novel non-backpropagation-based IFFCL algorithm.

\section{Conclusion}
$\text{VAE}$ is a type of autoencoder commonly used for data generation. In this paper, we propose a technique that, rather than treating it as a fully fledged deep learning model, views it as a neural network layer. The $\text{IFFCL}$ algorithm a non-backpropagation based method was employed to train models constructed with this novel layer.

The results indicate that the model built with $\text{VAE}$ based layers could learn meaningful representations from the data; however, its overall performance remained limited, with both training and testing accuracies reaching only modest levels. These results fall significantly short of the performance achieved by current state-of-the-art models on the same datasets. While the proposed layer successfully demonstrated learning capability, further research is needed to address its limitations and enhance its overall effectiveness.

\printbibliography
\end{document}